\documentclass{article}

\usepackage{PRIMEarxiv}

\usepackage[utf8]{inputenc} 
\usepackage[T1]{fontenc}    
\usepackage{url}            
\usepackage{booktabs}       
\usepackage{amsfonts}       
\usepackage{nicefrac}       
\usepackage{microtype}      
\usepackage{lipsum}
\usepackage{fancyhdr}       
\usepackage{graphicx}       
\graphicspath{{media/}} 
\usepackage{amssymb}
\usepackage{colortbl}
\usepackage[colorlinks=true, urlcolor=blue]{hyperref}
\usepackage{lineno}
\usepackage{amsmath}
\usepackage{graphicx}
\usepackage{multirow}
\usepackage{booktabs}
\usepackage{xcolor}
\usepackage{xurl}

\providecommand{\mstd}[2]{\ensuremath{#1\,\pm\,#2}}
\definecolor{TableHeader}{HTML}{EEF3F8}
\newcommand{\tableheadershade}{\rowcolor{TableHeader}}
\newcommand{\tablesetup}{\setlength{\tabcolsep}{5pt}\renewcommand{\arraystretch}{1.12}}
\newcommand{\metricstablesetup}{\setlength{\tabcolsep}{3pt}\renewcommand{\arraystretch}{1.12}}

\title{TotalSynth: Robust Whole-Body Synthetic CT from MRI and CBCT
}

\author{
    Valentin Boussot$^{1\dagger}$, C{\'e}dric H{\'e}mon$^{1\dagger}$, Anaïs Barateau$^{1}$, Caroline Lafond$^{1}$, Jean-Claude Nunes$^{1}$, Jean-Louis Dillenseger$^{1}$\\
    $^1$ Univ. Rennes, CLCC Eug\`ene Marquis, INSERM, LTSI - UMR 1099\\
    F-35000 Rennes, France\\
    $^\dagger$ These authors contributed equally to this work.\\
}

\date{}

\begin{document}
\maketitle
\begin{abstract}
\sloppy
\textbf{Purpose:} To develop and evaluate TotalSynth, a reusable pretrained framework for whole-body synthetic CT (sCT) generation from MRI and cone-beam CT (CBCT) images. \\
\textbf{Materials and Methods:} In this retrospective technical study, the dataset was assembled between 2020 and 2026 from SynthRAD challenge data, four prostate cohorts, and BIC-MAC. After registration quality control, 1450 of 1800 public challenge pairs were retained; 350 were excluded for insufficient registration quality or major source/CT mismatch. The corpus also included 84 additional prostate MRI/CT and 60 external BIC-MAC MRI/CT cases. Three 5-fold model families were evaluated with image-domain, anatomy-aware, registration-based, and uncertainty metrics. Age and sex were not consistently available across public datasets.\\
\textbf{Results:} The released MRI-to-CT model achieved an overall MAE of 67.49 HU, SSIM of 0.920, and PSNR of 29.28 dB. The CBCT-to-CT model achieved an overall MAE of 53.55 HU, SSIM of 0.939, and PSNR of 32.09 dB. The unified model maintained similar performance on MRI inputs (MAE, 67.68 HU) and CBCT inputs (MAE, 54.22 HU). On external BIC-MAC data, MRI-to-CT MAE was 100.91 HU without fine-tuning and 62.21 HU after fine-tuning.\\
\textbf{Conclusion:} TotalSynth provides reusable MRI- and CBCT-based CT synthesis models with broad anatomical coverage, while external evaluation highlights the need for local validation and optional fine-tuning under domain shift.
\end{abstract}

\section*{Summary Statement}
TotalSynth provides reusable computed tomography synthesis models from magnetic resonance imaging and cone-beam computed tomography, with broad anatomical coverage and improved external performance after fine-tuning.

\section*{Key Results}
\begin{itemize}
    \item The training corpus combined public challenge data, four additional prostate MRI cohorts, and external whole-body MRI data for registered CT synthesis supervision.
    \item Dedicated MRI-to-CT and cone-beam CT-to-CT models achieved overall mean absolute errors of 67.49 HU and 53.55 HU, respectively.
    \item On external whole-body MRI data, fine-tuning reduced mean absolute error from 100.91 HU to 62.21 HU.
\end{itemize}

\section{Introduction}

Computed tomography (CT) remains central to radiotherapy because it provides both patient geometry and Hounsfield unit (HU) values for electron density assignment and dose calculation. Yet many current workflows rely increasingly on magnetic resonance imaging (MRI), which offers superior soft-tissue contrast for delineation and MRI-guided radiotherapy, or on cone-beam CT (CBCT), which provides repeated volumetric imaging for positioning and adaptive treatment \cite{bahloul2024advancements,boulanger2021deep}. In both settings, missing or unreliable CT-like attenuation information limits dose-related analyses, registration, segmentation, and quantitative downstream tasks.

Synthetic CT (sCT) generation addresses this limitation by estimating CT-equivalent images from MRI or CBCT. MRI-derived sCT can support MRI-only planning, while CBCT-derived sCT can reduce the impact of scatter, artifacts, truncated fields of view, and inaccurate HU values in adaptive radiotherapy \cite{liu2023review}. Deep learning has become the dominant paradigm for this task \cite{edmund2017review,spadea2021deep}, but many published models remain restricted to a single anatomy, modality, institution, or acquisition protocol. This specialization limits reuse in clinical research environments where scanner vendors, MRI contrasts, CBCT protocols, field of view, and anatomy vary substantially.

Recent SynthRAD2023 and SynthRAD2025 challenges have improved access to paired MRI/CT and CBCT/CT data across multiple anatomical regions \cite{huijben2024generating,thummerer2025synthrad2025}. However, the field still lacks reusable pretrained synthesis resources analogous in spirit to broadly applicable segmentation tools such as TotalSegmentator \cite{wasserthal2023totalsegmentator}. Such a resource should provide not only benchmark performance, but also curated supervision, released weights, external validation, and a practical deployment pathway.

Inspired by this philosophy, we introduce TotalSynth, a family of pretrained models for whole-body CT synthesis from MRI, CBCT, or both modalities. The contribution is not new raw image acquisition, but the harmonization of existing paired datasets into registered CT supervision and the release of reusable model families. Specifically:
\begin{enumerate}
    \item we assemble a multi-dataset corpus for MRI-to-CT and CBCT-to-CT synthesis with CT targets aligned to the source images;
    \item we train dedicated MRI-to-CT, dedicated CBCT-to-CT, and unified MRI/CBCT-to-CT 5-fold model families;
    \item we evaluate performance across anatomical regions, including external whole-body MRI data and fine-tuning under domain shift.
\end{enumerate}

\section{Materials and Methods}

\subsection{Study design and datasets}

\begin{figure}[h!]
\centering
\includegraphics[width=1\textwidth]{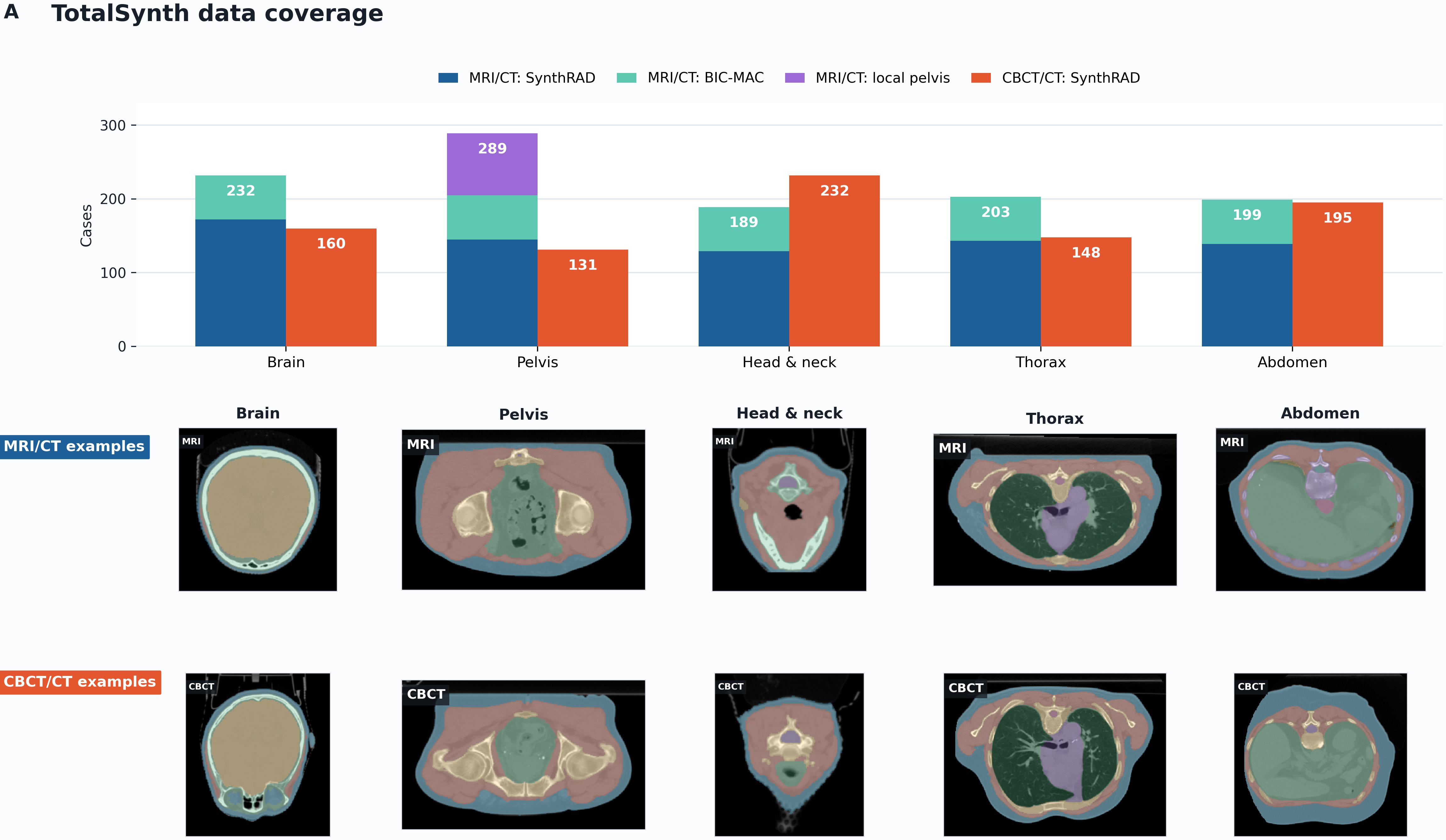}
\caption{TotalSynth data coverage by source modality and anatomical region. MRI = magnetic resonance imaging; CT = computed tomography; CBCT = cone-beam CT; BIC-MAC = Big Cross-Modal Attenuation Correction. The bar chart summarizes registered SynthRAD2023/2025 cases counted from the public registration repositories, 60 BIC-MAC whole-body MRI cases counted for each anatomical region, and 84 additional prostate MRI/CT cases. Representative aligned CT slices are shown with segmentation-mask overlays to illustrate anatomical coverage and registration-derived supervision.}
\label{fig:dataset_atlas}
\end{figure}

This retrospective technical study used paired MRI/CT and CBCT/CT data from public radiotherapy challenges, complementary prostate MRI cohorts, and external whole-body MRI data (Table~\ref{tab:datasets}; Figure~\ref{fig:dataset_atlas}). SynthRAD2023 provided brain and pelvic MRI-to-CT and CBCT-to-CT cases from three Dutch university medical centers \cite{thummerer2023synthrad2023,huijben2024generating}. SynthRAD2025 extended the corpus to head-and-neck, thoracic, and abdominal sites, with MRI/CT and CBCT/CT pairs collected across five European university hospitals \cite{thummerer2025synthrad2025,rogowski2026generating}. Together, these challenge datasets contributed broad variability in anatomy, scanner vendor, acquisition protocol, field of view, and image quality.

The public challenge data were deliberately heterogeneous. SynthRAD2023 included brain and pelvic cases acquired for photon and proton radiotherapy workflows, with MRI contrasts spanning T1-weighted gradient echo, inversion recovery, spoiled gradient echo, and T2-weighted sequences acquired at 1.5T and 3T. SynthRAD2025 further broadened the acquisition domain with head-and-neck, thoracic, and abdominal cases, including low-field MR-Linac, diagnostic MRI, and CBCT acquisitions from multiple image-guided radiotherapy systems. This heterogeneity was central to the study design because the intended use of TotalSynth is cross-institutional reuse rather than optimization for a single scanner or anatomy.

Four additional prostate cohorts (D1--D4) enriched pelvic MRI-to-CT training with treatment-position CT and MRI, including conventional diagnostic MRI, GoldAtlas data \cite{nyholm2018mr}, and MRI-Linac acquisitions. Dataset D1 included 24 patients with 3D T2-weighted SPACE MRI \cite{dowling2015automatic}; D2 included 18 patients with T2-weighted MRI and bladder contrast at CT; D3 corresponded to 17 GoldAtlas patients; and D4 included 25 patients imaged on a 1.5T Unity MRI-Linac.

The Big Cross-Modal Attenuation Correction (BIC-MAC) dataset \cite{hinge2026bicmac} was used only for fine-tuning and out-of-distribution evaluation. We used 60 whole-body paired MR Dixon/CT cases from healthy volunteers, acquired from vertex to mid-thigh. An additional external CSIRO MRI-to-CT subset was retained for MRI-domain evaluation and reported separately in the regional results.

BIC-MAC was intentionally kept outside the primary training corpus to test domain shift. Its whole-body Dixon MRI and low-dose attenuation-correction CT protocol differs from the radiotherapy challenge data in anatomy, contrast mechanism, subject population, and scan coverage, making it a useful stress test for direct generalization and local fine-tuning.

\subsection{Case selection and quality control}

The study dataset was assembled between 2020 and 2026. Public SynthRAD cases were eligible when the source image and reference CT were available and when deformable registration produced an anatomically plausible CT target in the source-image geometry. Cases were excluded when registration quality was insufficient, when the secondary image and original CT showed major anatomical or field-of-view differences, or when the registered supervision could not be visually accepted for model training or evaluation. Based on the original challenge sizes, 618 of 720 SynthRAD2023 cases and 832 of 1080 SynthRAD2025 cases were retained after quality control; 350 patients were excluded. The four additional prostate cohorts contributed 84 MRI/CT cases after eligibility screening, and 60 BIC-MAC cases were used for external evaluation and fine-tuning. No formal sample-size calculation was performed; all eligible cases were used. Age and sex were not consistently available across all public challenge datasets and were therefore not analyzed.

\begin{table}[!t]
\centering
\caption{Summary of datasets used in this study after eligibility screening and registration quality control. MRI = magnetic resonance imaging; CT = computed tomography; CBCT = cone-beam CT; OoD = out-of-distribution.}
\label{tab:datasets}
\small
\tablesetup
\resizebox{\textwidth}{!}{%
\begin{tabular}{@{}llllrrl@{}}
\toprule
\tableheadershade Dataset & Modality pair & Region & Source cases & Retained cases & Excluded cases & Use \\
\midrule
SynthRAD2023 & MRI+CBCT & Brain, Pelvis & 720 & 618 & 102 & Training/evaluation \\
SynthRAD2025 & MRI+CBCT & Head \& Neck, Thorax, Abdomen & 1080 & 832 & 248 & Training/evaluation \\
BIC-MAC & MRI(Dixon)/CT & Whole-body & 75 & 60 & 15 & Fine-tuning/OoD \\
D1 & MRI/CT & Prostate & 24 & 24 & 0 & Training/evaluation \\
D2 & MRI/CT & Prostate & 18 & 18 & 0 & Training/evaluation \\
GoldAtlas (D3) & MRI/CT & Prostate & 19 & 17 & 2 & Training/evaluation \\
D4 & MRI/CT & Prostate & 25 & 25 & 0 & Training/evaluation \\
\bottomrule
\end{tabular}}
\end{table}

\subsection{CT target generation}

For supervised training, the planning CT was spatially aligned to each source image: CT to MRI for MRI-to-CT training and CT to CBCT for CBCT-to-CT training. The registered CT served as the target image, while the MRI or CBCT image served as the model input. Registration was part of dataset curation and was not required during released-model inference. Consequently, the target should be interpreted as a quality-controlled registered CT reference rather than a perfect anatomical ground truth, particularly in cases with organ motion, truncation, artifacts, or large inter-scan anatomical differences.

Registrations were generated with the IMPACT framework \cite{boussot2025impact} using deformable multimodal registration informed by semantic anatomical features. Each deformation field was visually reviewed, and cases with implausible deformation, missing anatomy, severe field-of-view mismatch, or unresolved inter-scan anatomical change were excluded before training. This quality-control step was treated as part of the dataset annotation process. It also makes the supervision bias explicit: the model is optimized against accepted registered CT targets, and residual registration error remains a measurable limitation of the training signal.

\subsection{TotalSynth pipeline overview}

\begin{figure}[h!]
\centering
\includegraphics[width=1\textwidth]{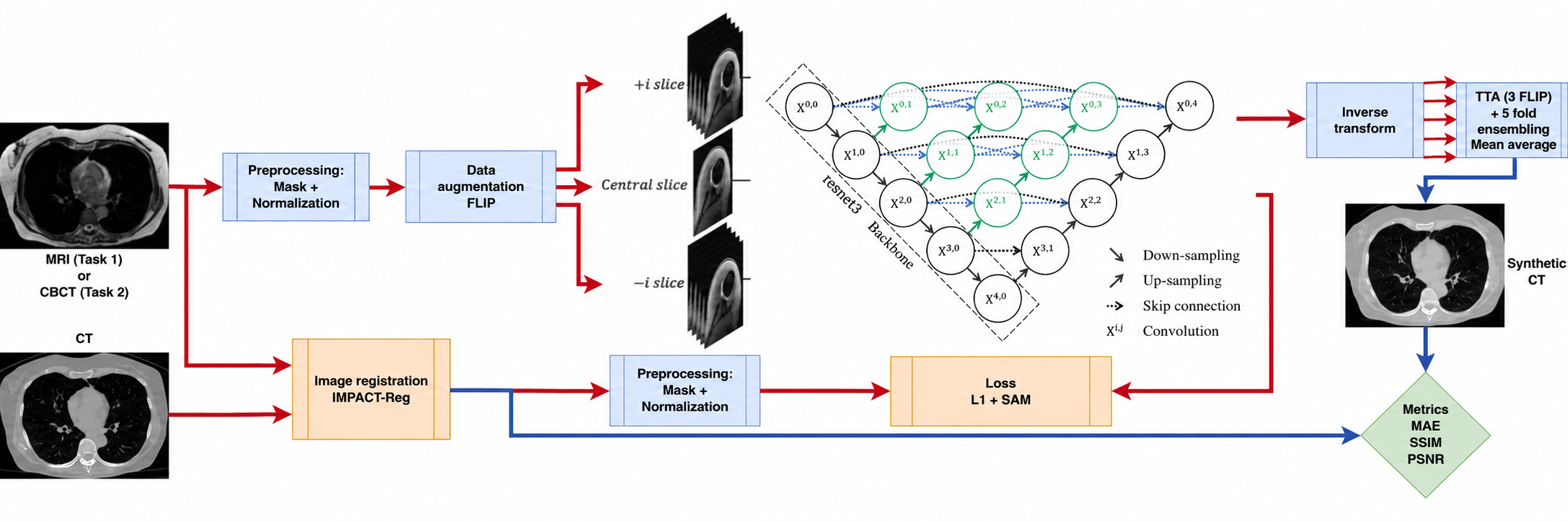}
\caption{Overview of the TotalSynth training and inference pipeline. MRI = magnetic resonance imaging; CT = computed tomography; CBCT = cone-beam CT. MRI or CBCT images are preprocessed and used as input to a 2.5D synthesis model that predicts a CT-like image. During training, reference CT images are spatially aligned to the input images and used as supervised targets. During inference, predictions are generated using test-time augmentation and 5-fold model ensembling, then restored to the original image geometry.}
\label{fig:pipeline}
\end{figure}

TotalSynth uses a common preprocessing, training, and inference framework for MRI and CBCT inputs. Three 5-fold pretrained model families are released: a dedicated MRI-to-CT model, a dedicated CBCT-to-CT model, and a unified MRI/CBCT-to-CT model that accepts both source modalities.

The 5-fold design was used for two reasons. First, it allowed all training cases to contribute to model development while preserving held-out evaluation within each fold. Splits were defined at the patient or case level so that a source image and its registered CT target were never used simultaneously for training and held-out testing within a fold. Second, it enabled ensemble prediction at inference time, which improved stability across heterogeneous inputs and provided a simple uncertainty estimate from inter-model and test-time augmentation variability. All quantitative results reported for the primary evaluation were computed on a held-out set of cases that was not used to train any of the five ensemble members, and every member was applied to the same cases.

\subsection{Preprocessing}

Preprocessing used a body mask to restrict synthesis to the relevant anatomy, suppress background, and exclude regions where the registered CT extended beyond the source-image field of view. In the released inference pipeline, body masks are generated with the IMPACTSeg body model, resampled with the input image to 1 $\times$ 1 $\times$ 3 mm spacing, and dilated by five voxels to preserve the skin surface and nearby anatomy. CT targets were clipped to $[-1024, 3071]$ HU and rescaled to $[-1, 1]$. MRI and CBCT inputs were clipped between their minimum intensity and the $99.5^{\mathrm{th}}$ percentile, then rescaled to $[-1, 1]$ using image-specific intensity ranges. Images are processed as axial 512 $\times$ 512 patches with neighboring-slice context, and voxels outside the body mask are assigned air intensity in the final sCT.

\subsection{2.5D synthesis model}

TotalSynth uses a 2.5D strategy in which five adjacent slices are provided as input and the network predicts the CT corresponding to the central slice. This design keeps memory requirements compatible with whole-volume inference while giving the network limited through-plane context, which is important for thin structures, air cavities, and anatomy affected by anisotropic slice spacing. The generator is a U-Net++ with a ResNet-34 encoder \cite{zhou1807nested,he2016deep}, 26 084 881 trainable parameters, and slice-wise inference over the full volume.

\subsection{Training of the released models}

The dedicated MRI-to-CT and CBCT-to-CT models were trained on their respective paired datasets, whereas the unified model used both MRI/CT and CBCT/CT pairs. Each model family was trained as a five-fold ensemble. Random augmentation was used during training, and the objective combined voxel-wise agreement with the registered CT target and an anatomical consistency term.

\subsection{Inference}

At inference, the input image is preprocessed as during training, synthesized slice by slice by each fold, and combined with flipped test-time augmentations. Predictions are averaged across augmentations and ensemble members, then restored to the original image geometry to produce an sCT aligned with the input MRI or CBCT image. Voxel-wise dispersion across ensemble members and test-time augmentations was retained as an uncertainty map, allowing high-variability regions to be inspected alongside the synthesized image.

\subsection{Evaluation metrics}

All metrics were computed inside the body mask used for synthesis. Let $\hat{y}$ denote the sCT, $y$ the registered CT target and $x$ the source MRI or CBCT image.

\textbf{MAE, PSNR, SSIM} quantify voxel-wise and structural agreement between $\hat{y}$ and $y$, with the dynamic range taken from each image pair rather than fixed a priori.

\textbf{Dice} compares TotalSegmentator \cite{wasserthal2023totalsegmentator} segmentations of $\hat{y}$ and $y$. Per-label values are averaged within each case, then across cases; since the labels present vary by region, the reported Dice is a mean over the structures actually present rather than a fixed anatomical panel.

\textbf{SAM} is a perceptual distance in the feature space of the frozen Segment Anything Model~2.1 encoder \cite{ravi2025sam2}, restricted to its two intermediate levels:
\begin{equation}
d_{\mathrm{SAM}}(\hat{y},y) \;=\; \sum_{l}\sum_{c} \alpha_{l,c}\,\bigl\| \phi_{l,c}(\hat{y}) - \phi_{l,c}(y) \bigr\|_1 ,
\end{equation}
where $\phi_{l,c}$ is feature channel $c$ at level $l$ and $\alpha_{l,c}\geq0$ are channel weights calibrated so that a registration-deformed CT control stays closer to the reference CT than an sCT trained with voxel-wise supervision alone.

\textbf{Reg} applies the IMPACT formulation \cite{boussot2025impact} to features of the frozen TotalSegmentator CT 3\,mm encoder on $128^3$ patches. Unlike every other metric here, it compares $\hat{y}$ to the source image $x$, and therefore measures how far synthesis has displaced the anatomy it was given; lower is better. An sCT close in HU but geometrically shifted relative to its own input is unusable for contour propagation or dose accumulation, which no reference-based metric detects.

\textbf{Uncertainty} is the mean voxel-wise variance (HU$^2$) of the predictions across the five ensemble members and the flipped test-time augmentations, and is therefore a lower bound on total predictive uncertainty.

\subsection{Statistical analysis and software}

All statistical analyses were descriptive. Metrics were summarized as mean $\pm$ standard deviation across cases and anatomical regions, and no formal hypothesis testing was performed. Model training, inference, and evaluation used the Python-based TotalSynth/KonfAI pipeline; released-model deployment was implemented in the IMPACT-Synth 3D Slicer extension \cite{fedorov2012slicer}. Image-domain and anatomy-aware metric calculations used the same software pipeline for all model families to ensure consistent comparisons.

\section{Results}

\begin{figure}[h!]
\centering
\includegraphics[width=1\textwidth]{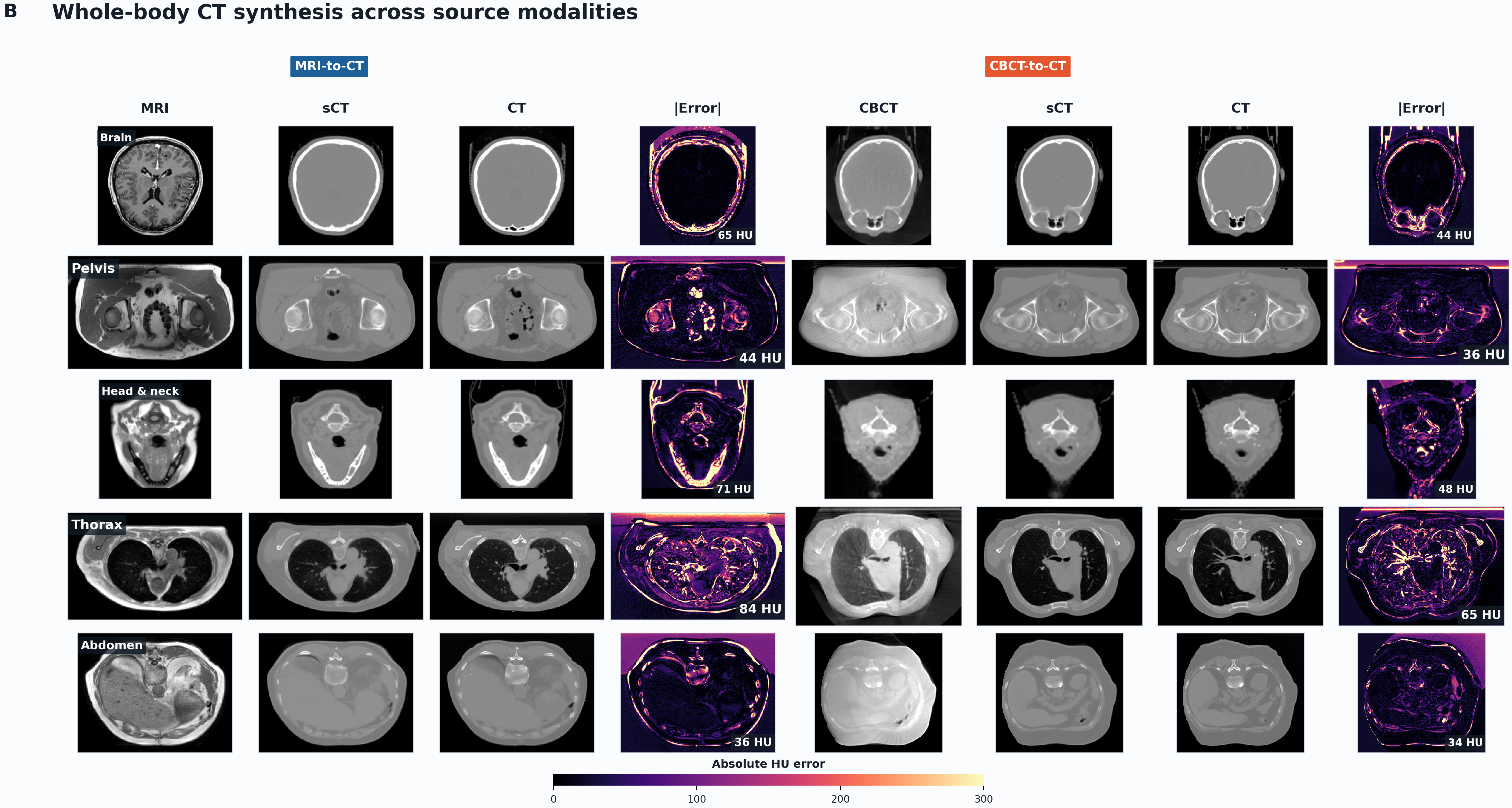}
\caption{Representative CT synthesis examples across source modalities and anatomical regions. MRI = magnetic resonance imaging; CT = computed tomography; CBCT = cone-beam CT; sCT = synthetic CT; HU = Hounsfield unit. For each region, the source MRI or CBCT image, generated sCT, registered CT target, and absolute HU error map are shown. These examples were generated with a single ensemble member and no test-time augmentation for visual illustration; quantitative tables report the released model evaluation.}
\label{fig:synthesis_capability}
\end{figure}

We evaluated the released dedicated MRI-to-CT model, dedicated CBCT-to-CT model, and unified MRI/CBCT-to-CT model. Representative synthesis examples are shown in Figure~\ref{fig:synthesis_capability}, and overall quantitative results are summarized in Table~\ref{tab:main_performance_summary}. Lower MAE, SAM, registration error, and uncertainty indicate better performance, whereas higher PSNR, SSIM, and Dice indicate better agreement with registered CT targets or downstream anatomical structures. Because the reference CT images are registration-derived, all metrics should be interpreted as agreement with curated supervision rather than absolute ground truth accuracy.

\subsection{Released model performance}

The dedicated MRI-to-CT model achieved an overall MAE of 67.49 HU, SSIM of 0.920, and PSNR of 29.28 dB across abdominal, head-and-neck, thoracic, brain, pelvic, and CSIRO cases (Table~\ref{tab:main_performance_summary}). Errors were lowest in the pelvis and higher in head-and-neck, brain, and CSIRO cases, consistent with the greater heterogeneity of MRI contrast, anatomy, and acquisition protocols in these domains (Appendix Table~\ref{tab:mri_to_ct_release_cv_metrics}).

The dedicated CBCT-to-CT model achieved an overall MAE of 53.55 HU, SSIM of 0.939, and PSNR of 32.09 dB. Performance was strongest in brain and pelvis and lower in head-and-neck, where anatomical complexity, field-of-view variation, and CBCT artifacts are more prominent (Appendix Table~\ref{tab:cbct_to_ct_release_cv_metrics}).

\begin{table}[!t]
\centering
\caption{Overall performance of released TotalSynth models and BIC-MAC fine-tuning. MRI = magnetic resonance imaging; CT = computed tomography; CBCT = cone-beam CT; BIC-MAC = Big Cross-Modal Attenuation Correction; MAE = mean absolute error; HU = Hounsfield unit; SSIM = structural similarity index; PSNR = peak signal-to-noise ratio. Values report mean $\pm$ standard deviation. Detailed regional results are provided in the Appendix.}
\label{tab:main_performance_summary}
\small
\tablesetup
\begin{tabular}{@{}llccc@{}}
\toprule
\tableheadershade
Model and setting & Evaluation domain & MAE (HU) & SSIM & PSNR (dB) \\
\midrule
MRI-to-CT & Primary MRI/CT & \mstd{67.49}{15.97} & \mstd{0.920}{0.025} & \mstd{29.28}{2.10} \\
CBCT-to-CT & Primary CBCT/CT & \mstd{53.55}{17.79} & \mstd{0.939}{0.026} & \mstd{32.09}{2.93} \\
Unified model & MRI inputs & \mstd{67.68}{16.42} & \mstd{0.920}{0.025} & \mstd{29.22}{2.18} \\
Unified model & CBCT inputs & \mstd{54.22}{17.54} & \mstd{0.938}{0.025} & \mstd{31.89}{2.87} \\
MRI-to-CT, direct & BIC-MAC & \mstd{100.91}{18.67} & \mstd{0.968}{0.007} & \mstd{26.69}{1.40} \\
MRI-to-CT, fine-tuned & BIC-MAC & \mstd{62.21}{6.19} & \mstd{0.982}{0.002} & \mstd{30.45}{0.92} \\
\bottomrule
\end{tabular}
\end{table}

\subsection{Unified model and external adaptation}

The unified MRI/CBCT-to-CT model maintained performance close to the modality-specific models, with overall MAE values of 67.68 HU on MRI inputs and 54.22 HU on CBCT inputs (Table~\ref{tab:main_performance_summary}; Appendix Table~\ref{tab:unified_mr_cbct_release_cv_metrics}). This supports the feasibility of a single reusable synthesis model when simplified deployment is preferred over separate modality-specific pipelines.

Each BIC-MAC subject was evaluated under both contrasts reconstructed from the same whole-body Dixon acquisition: IN denotes the in-phase image and OUT the out-of-phase image. Because the two derive from the same echoes, they share identical geometry and motion state, so the difference between them isolates the effect of input contrast from anatomy and registration. Fine-tuning and evaluation used disjoint subjects, 45 and 15 respectively, of the 60 retained cases.

On external BIC-MAC whole-body MRI data, direct inference increased MAE to 100.91 HU, indicating a substantial domain shift. Fine-tuning reduced MAE to 62.21 HU and increased PSNR from 26.69 dB to 30.45 dB, with improvements for both in-phase and out-of-phase inputs (Appendix Table~\ref{tab:bicmac_external_finetune}). These results support the use of TotalSynth both as a direct inference model and as an initialization for local adaptation.

\section{Discussion}

TotalSynth was designed as a reusable pretrained resource for radiotherapy sCT generation, rather than as a model optimized for a single anatomy or acquisition protocol. This positioning follows the broader trend toward deployable medical imaging tools such as TotalSegmentator \cite{wasserthal2023totalsegmentator}, but addresses cross-modality CT synthesis. The results indicate that multi-dataset harmonization, registered CT supervision, five-fold ensembling, and released weights can support MRI-to-CT and CBCT-to-CT synthesis across several anatomical regions.

The unified MRI/CBCT-to-CT model performed close to the dedicated modality-specific models. This is practically relevant because a single model can simplify deployment in centers exploring both MRI-only and CBCT-based adaptive workflows. Dedicated models may remain preferable when the input modality and target deployment setting are narrowly defined, whereas the unified model provides a broader starting point for validation and adaptation.

The BIC-MAC evaluation emphasizes the importance of external validation \cite{yu2022external}. Direct inference showed a clear MAE increase despite high SSIM, illustrating that structural similarity can coexist with clinically relevant HU deviations. This degradation is likely related to differences in whole-body coverage, Dixon MRI contrast, healthy-volunteer imaging, and CT protocol \cite{guan2021domain,guo2026impact}. Fine-tuning substantially reduced MAE, supporting the use of TotalSynth as an initialization model for local adaptation rather than as a frozen universal solution.

A key methodological feature is the use of registered CT targets. This enables supervised learning across datasets acquired at different time points and geometries, but also introduces supervision bias when registration is imperfect. Manual verification and exclusion of problematic deformation fields are therefore part of the contribution. The IMPACT-based deformation fields and metadata are available at \url{https://huggingface.co/datasets/VBoussot/synthrad2023-impact-registration} and \url{https://huggingface.co/datasets/VBoussot/synthrad2025-impact-registration}, making the registration-derived supervision inspectable rather than hidden.

Clinical translation will require task-specific quality assurance. The released 3D Slicer IMPACT-Synth plugin powered by KonfAI \cite{boussot2025konfai}, provides a deployment pathway for local inference, visual inspection, uncertainty visualization, and structure-based comparison. These tools are intended for research and validation workflows; dose calculation, contour propagation, attenuation correction, and adaptive decision-making each require specific acceptance criteria.

From a dissemination perspective, TotalSynth was structured to be inspectable and reusable rather than only reported as a trained network. The release separates three components that are often intertwined in sCT studies: curated registered supervision, pretrained synthesis weights, and a deployment interface. This separation allows external groups to audit the registration targets, run the same released models on local MRI or CBCT images, and determine whether direct inference is sufficient or whether center-specific fine-tuning is required. The BIC-MAC experiment illustrates this workflow: direct inference quantified domain shift, whereas fine-tuning converted the released MRI-to-CT model into a substantially stronger local model. In radiotherapy, where scanner protocols, immobilization, patient positioning, and downstream tolerances vary across institutions, this practical validation pathway may be as important as average image-domain performance. It also establishes a common baseline for future studies of dose calculation, contour propagation, attenuation correction, adaptive radiotherapy quality assurance, and uncertainty-guided review.

Several limitations remain. Performance depends on the diversity and quality of the training data, and rare anatomy, unusual MRI contrasts, metal artifacts, severe truncation, and large anatomical changes remain challenging. Image-domain metrics do not fully characterize clinical usefulness, and ensemble-based uncertainty captures only part of the model uncertainty. Future work should expand the corpus, improve domain adaptation, explore registration-free or weakly supervised training, and quantify downstream dose and adaptive radiotherapy impact.

\section{Conclusion}

TotalSynth provides reusable pretrained models for whole-body synthetic CT generation from MRI and CBCT images. By combining harmonized multi-dataset supervision, quality-controlled registered CT targets, dedicated and unified model families, external evaluation, and open deployment resources, TotalSynth offers a practical baseline for MRI-only radiotherapy, CBCT-based adaptive radiotherapy, and CT-dependent downstream research. External BIC-MAC results show that domain shift remains important, but fine-tuning from released weights can substantially improve performance, supporting TotalSynth as both a direct inference tool and an initialization for local adaptation.

\section{Ethical Approval}
\label{sec:ethics}
The institutional cohorts D1, D2, and D4 were conducted in accordance with statutory requirements and approved by the regional research ethics committee (Comité de Protection des Personnes, Ouest V, Rennes, ARCOL 2015-02-45-01). All participants in these institutional cohorts gave written informed consent to participate in the study. GoldAtlas (D3) is a public dataset collected under its original ethical approval (dnr:2015/09-31), with informed consent from all included patients. The remaining public datasets were used according to their respective data access conditions and citation requirements.

\section{Funding}
\label{sec:acknowledgments}
The work presented in this article was supported by the French National Research Agency as part of the VATSop project (ANR-20-CE19-0015) and by the French National Research Agency as part of the DIMADOSE project (C. Hémon).

\section{Disclosures}
The authors have no relevant financial or non-financial interests to disclose.

\section{Use of Artificial Intelligence Tools}
While preparing this work, the authors used ChatGPT to enhance the writing structure and refine grammar. After using these tools, the authors reviewed and edited the content as needed and take full responsibility for the publication's content.

\section{Data and Code Availability}
The registration-derived SynthRAD2023 and SynthRAD2025 supervision, deformation fields, and associated metadata are available at \url{https://huggingface.co/datasets/VBoussot/synthrad2023-impact-registration} and \url{https://huggingface.co/datasets/VBoussot/synthrad2025-impact-registration}. Access to the original public challenge images remains governed by the corresponding SynthRAD data access conditions. The released TotalSynth inference workflow is available through the IMPACT-Synth 3D Slicer extension at \url{https://github.com/vboussot/SlicerImpactSynth} and associated KonfAI model repositories. Institutional prostate cohorts cannot be publicly redistributed because of consent and privacy restrictions.

\bibliographystyle{unsrt}  
\bibliography{biblio.bib}

\appendix

\clearpage
\section*{Appendix: Detailed Quantitative Results}

\noindent\textit{Abbreviations used in the Appendix tables:} MAE = mean absolute error; PSNR = peak signal-to-noise ratio; SSIM = structural similarity index; SAM = Segment Anything Model-based anatomical metric; Reg = registration-based score.

\begin{table}[!ht]
\centering
\caption{Evaluation of the released MRI-to-CT model by anatomical region. Values report mean $\pm$ standard deviation. Case counts refer to the image-domain and Dice evaluation.}
\label{tab:mri_to_ct_release_cv_metrics}
\scriptsize
\metricstablesetup
\begin{tabular}{@{}lccccccc@{}}
\toprule
\tableheadershade
Region & MAE & PSNR & SSIM & Dice & SAM & Reg & Uncertainty \\
\midrule
AB ($n=22$) & \mstd{60.10}{9.42} & \mstd{30.01}{1.47} & \mstd{0.908}{0.022} & \mstd{0.598}{0.070} & \mstd{27.405}{4.156} & \mstd{0.511}{0.117} & \mstd{1543.81}{580.52} \\
HN ($n=21$) & \mstd{80.89}{13.12} & \mstd{28.44}{1.35} & \mstd{0.925}{0.030} & \mstd{0.645}{0.106} & \mstd{13.412}{2.108} & \mstd{0.435}{0.066} & \mstd{1308.83}{512.26} \\
TH ($n=23$) & \mstd{58.45}{11.20} & \mstd{31.21}{1.86} & \mstd{0.940}{0.019} & \mstd{0.645}{0.051} & \mstd{21.638}{4.232} & \mstd{0.466}{0.089} & \mstd{751.61}{262.12} \\
brain ($n=27$) & \mstd{79.71}{8.64} & \mstd{27.44}{0.92} & \mstd{0.915}{0.015} & \mstd{0.751}{0.096} & \mstd{13.209}{2.612} & \mstd{1.199}{0.138} & \mstd{3402.08}{633.51} \\
pelvis ($n=23$) & \mstd{48.16}{8.30} & \mstd{31.28}{1.42} & \mstd{0.901}{0.025} & \mstd{0.756}{0.108} & \mstd{25.949}{3.617} & \mstd{0.574}{0.094} & \mstd{2348.75}{648.73} \\
CSIRO ($n=24$)  & \mstd{75.99}{10.13} & \mstd{27.67}{1.02} & \mstd{0.929}{0.016} & \mstd{0.774}{0.035} & \mstd{30.898}{3.947} & \mstd{0.434}{0.054} & \mstd{1272.28}{153.55} \\
\midrule
overall ($n=140$) & \mstd{67.49}{15.97} & \mstd{29.28}{2.10} & \mstd{0.920}{0.025} & \mstd{0.698}{0.106} & \mstd{21.981}{7.692} & \mstd{0.622}{0.303} & \mstd{1822.49}{1035.84} \\
\bottomrule
\end{tabular}
\end{table}

\begin{table}[!ht]
\centering
\caption{Evaluation of the released CBCT-to-CT model by anatomical region. Values report mean $\pm$ standard deviation.}
\label{tab:cbct_to_ct_release_cv_metrics}
\scriptsize
\metricstablesetup
\begin{tabular}{@{}lccccccc@{}}
\toprule
\tableheadershade
Region & MAE & PSNR & SSIM & Dice & SAM & Reg & Uncertainty \\
\midrule
AB ($n=32$) & \mstd{57.18}{14.96} & \mstd{31.12}{2.42} & \mstd{0.926}{0.023} & \mstd{0.673}{0.086} & \mstd{17.706}{4.475} & \mstd{1.049}{0.016} & \mstd{730.95}{198.42} \\
HN ($n=37$) & \mstd{67.53}{15.80} & \mstd{29.97}{2.19} & \mstd{0.946}{0.021} & \mstd{0.717}{0.075} & \mstd{8.853}{1.528} & \mstd{1.052}{0.036} & \mstd{661.14}{231.48} \\
TH ($n=34$) & \mstd{57.27}{15.33} & \mstd{32.03}{2.69} & \mstd{0.928}{0.019} & \mstd{0.724}{0.075} & \mstd{16.241}{4.455} & \mstd{1.045}{0.021} & \mstd{793.57}{252.09} \\
brain ($n=26$) & \mstd{39.42}{6.13} & \mstd{33.96}{1.56} & \mstd{0.970}{0.007} & \mstd{0.905}{0.071} & \mstd{7.219}{1.024} & \mstd{1.013}{0.002} & \mstd{879.35}{256.72} \\
pelvis ($n=21$) & \mstd{34.84}{8.32} & \mstd{35.12}{2.49} & \mstd{0.924}{0.023} & \mstd{0.835}{0.078} & \mstd{20.066}{4.641} & \mstd{1.023}{0.005} & \mstd{949.44}{381.04} \\
\midrule
overall ($n=150$) & \mstd{53.55}{17.79} & \mstd{32.09}{2.93} & \mstd{0.939}{0.026} & \mstd{0.758}{0.113} & \mstd{13.703}{6.015} & \mstd{1.039}{0.027} & \mstd{784.24}{278.43} \\
\bottomrule
\end{tabular}
\end{table}

\begin{table}[!ht]
\centering
\caption{Evaluation of the released unified MRI/CBCT-to-CT model by input domain and anatomical region. Values report mean $\pm$ standard deviation.}
\label{tab:unified_mr_cbct_release_cv_metrics}
\scriptsize
\metricstablesetup
\begin{tabular}{@{}lccccccc@{}}
\toprule
\tableheadershade
Region & MAE & PSNR & SSIM & Dice & SAM & Reg & Uncertainty \\
\midrule
MRI-to-CT/AB & \mstd{59.64}{9.56} & \mstd{30.04}{1.48} & \mstd{0.909}{0.023} & \mstd{0.591}{0.066} & \mstd{27.362}{4.014} & \mstd{0.526}{0.140} & \mstd{1451.52}{614.79} \\
MRI-to-CT/HN & \mstd{80.34}{12.50} & \mstd{28.39}{1.34} & \mstd{0.926}{0.029} & \mstd{0.641}{0.106} & \mstd{13.378}{2.122} & \mstd{0.423}{0.080} & \mstd{1181.19}{467.90} \\
MRI-to-CT/TH & \mstd{58.00}{11.26} & \mstd{31.19}{1.88} & \mstd{0.940}{0.019} & \mstd{0.635}{0.054} & \mstd{21.851}{4.555} & \mstd{0.436}{0.087} & \mstd{673.25}{226.95} \\
MRI-to-CT/brain & \mstd{80.76}{9.21} & \mstd{27.26}{0.99} & \mstd{0.913}{0.016} & \mstd{0.763}{0.108} & \mstd{13.355}{2.870} & \mstd{1.185}{0.132} & \mstd{2897.51}{530.13} \\
MRI-to-CT/pelvis & \mstd{47.43}{8.03} & \mstd{31.35}{1.49} & \mstd{0.902}{0.025} & \mstd{0.771}{0.102} & \mstd{27.045}{3.892} & \mstd{0.579}{0.096} & \mstd{2171.19}{487.25} \\
MRI-to-CT/CSIRO & \mstd{76.98}{10.54} & \mstd{27.55}{1.05} & \mstd{0.928}{0.016} & \mstd{0.775}{0.038} & \mstd{31.038}{4.174} & \mstd{0.274}{0.060} & \mstd{1303.97}{165.31} \\
MRI-to-CT/overall & \mstd{67.68}{16.42} & \mstd{29.22}{2.18} & \mstd{0.920}{0.025} & \mstd{0.701}{0.112} & \mstd{22.190}{7.884} & \mstd{0.585}{0.324} & \mstd{1644.98}{871.03} \\
\midrule
CBCT-to-CT/AB & \mstd{56.79}{14.26} & \mstd{31.07}{2.40} & \mstd{0.927}{0.024} & \mstd{0.665}{0.084} & \mstd{18.896}{4.771} & \mstd{0.013}{0.004} & \mstd{582.34}{212.75} \\
CBCT-to-CT/HN & \mstd{67.95}{15.63} & \mstd{29.86}{2.16} & \mstd{0.947}{0.021} & \mstd{0.707}{0.076} & \mstd{10.396}{1.583} & \mstd{0.012}{0.004} & \mstd{479.47}{162.48} \\
CBCT-to-CT/TH & \mstd{58.41}{14.72} & \mstd{31.76}{2.56} & \mstd{0.926}{0.018} & \mstd{0.726}{0.061} & \mstd{17.755}{4.460} & \mstd{0.015}{0.004} & \mstd{648.19}{208.79} \\
CBCT-to-CT/brain & \mstd{40.59}{6.22} & \mstd{33.51}{1.48} & \mstd{0.968}{0.007} & \mstd{0.907}{0.070} & \mstd{9.425}{2.031} & \mstd{0.028}{0.011} & \mstd{821.30}{268.82} \\
CBCT-to-CT/pelvis & \mstd{34.57}{7.72} & \mstd{35.19}{2.48} & \mstd{0.925}{0.022} & \mstd{0.828}{0.081} & \mstd{22.728}{4.836} & \mstd{0.023}{0.008} & \mstd{829.13}{292.40} \\
CBCT-to-CT/overall & \mstd{54.22}{17.54} & \mstd{31.89}{2.87} & \mstd{0.938}{0.025} & \mstd{0.752}{0.113} & \mstd{15.420}{6.138} & \mstd{0.017}{0.009} & \mstd{644.19}{260.67} \\
\bottomrule
\end{tabular}
\end{table}

\begin{table}[!ht]
\centering
\caption{BIC-MAC external evaluation before and after fine-tuning. Values report mean $\pm$ standard deviation.}
\label{tab:bicmac_external_finetune}
\scriptsize
\metricstablesetup
\begin{tabular}{@{}llccccccc@{}}
\toprule
\tableheadershade
Setting & Contrast & MAE & PSNR & SSIM & Dice & SAM & Reg & Uncertainty \\
\midrule
Direct & IN & \mstd{85.27}{6.15} & \mstd{27.89}{0.61} & \mstd{0.973}{0.003} & \mstd{0.568}{0.046} & \mstd{18.991}{1.608} & \mstd{0.180}{0.028} & \mstd{917.41}{115.51} \\
Direct & OUT & \mstd{115.50}{14.04} & \mstd{25.56}{0.91} & \mstd{0.962}{0.004} & \mstd{0.350}{0.070} & \mstd{20.804}{1.503} & \mstd{0.204}{0.027} & \mstd{2132.74}{263.72} \\
Direct & Overall & \mstd{100.91}{18.67} & \mstd{26.69}{1.40} & \mstd{0.968}{0.007} & \mstd{0.455}{0.124} & \mstd{19.929}{1.799} & \mstd{0.193}{0.030} & \mstd{1546.03}{641.28} \\
\midrule
Fine-tuned & IN & \mstd{60.00}{5.43} & \mstd{30.73}{0.87} & \mstd{0.983}{0.002} & \mstd{0.662}{0.025} & \mstd{13.113}{1.069} & \mstd{0.198}{0.028} & \mstd{272.78}{33.12} \\
Fine-tuned & OUT & \mstd{64.43}{6.12} & \mstd{30.16}{0.87} & \mstd{0.981}{0.002} & \mstd{0.648}{0.026} & \mstd{13.354}{1.063} & \mstd{0.209}{0.029} & \mstd{332.65}{40.43} \\
Fine-tuned & Overall & \mstd{62.21}{6.19} & \mstd{30.45}{0.92} & \mstd{0.982}{0.002} & \mstd{0.655}{0.026} & \mstd{13.234}{1.073} & \mstd{0.203}{0.029} & \mstd{302.72}{47.56} \\
\bottomrule
\end{tabular}
\end{table}

\end{document}